\pdfoutput=1

\documentclass[11pt]{article}
 
\usepackage[preprint]{acl}
 
\usepackage{times}
\usepackage{latexsym}
\usepackage{amsmath}
\usepackage{amssymb}
\usepackage{booktabs}

\usepackage[T1]{fontenc}

\usepackage[utf8]{inputenc}

\usepackage{microtype}

\usepackage{inconsolata}

\usepackage{graphicx}
\usepackage{commenting}

\usepackage{algorithm}
\usepackage{algpseudocode}
\usepackage{graphicx}
\usepackage{makecell}
\usepackage{fontawesome}
\usepackage[most]{tcolorbox}
\tcbset{
  mycompactbox/.style={
    colback=gray!10,
    colframe=black,
    boxsep=1pt,
    left=1pt,
    right=1pt,
    top=1pt,
    bottom=1pt,
    before skip=3pt,
    after skip=3pt,
    width=\columnwidth,
    center title
  }
}

\title{Final Checkpoints Are Not Enough: Analyzing Latent Reasoning Faithfulness Along Training Trajectories}

\author{
  \textbf{Hengyu Jin\textsuperscript{1,2,3}},
  \textbf{Shu Yang\textsuperscript{2,3}},
  \textbf{Di Wang\textsuperscript{2,3,*}}
  \\
  \textsuperscript{1}Tongji University \\
  \textsuperscript{2}Provable Responsible AI and Data Analytics (PRADA) Lab \\
  \textsuperscript{3}King Abdullah University of Science and Technology \\
  \normalsize\textsuperscript{*}Corresponding author \\
}

\begin{document}

\maketitle

\begin{abstract}
Latent reasoning performs multi-step inference in continuous hidden states, promising more compact and efficient reasoning. However, these opaque states raise a question of faithfulness: whether the latent reasoning steps drive the final answer. Prior work reports unfaithful behaviors at selected checkpoints, leaving their development during training unexamined. We track behavioral and activation-based evidence across training using verified counterfactual edits and interventions on the latent reasoning states. We find that accuracy can improve while models become less responsive to edits that change the required reasoning path and the correct answer. This counterfactual responsiveness follows distinct trajectories across latent reasoning approaches. On the graph reachability task ProsQA, output sensitivity to noise replacement declines alongside counterfactual responsiveness. Across separately trained binary-choice and open-ended versions of GSM8K math problems, interventions on all latent states show opposite sensitivity trajectories. Guided by these trajectories, we intervene on the training procedure and improve counterfactual responsiveness. These results show that evaluating only a final checkpoint can obscure both when counterfactual responsiveness changes and what the latent states contribute.
\end{abstract}

\section{Introduction}
\label{sec:intro}

\begin{figure*}[t]
\centering
\includegraphics[width=0.98\linewidth]{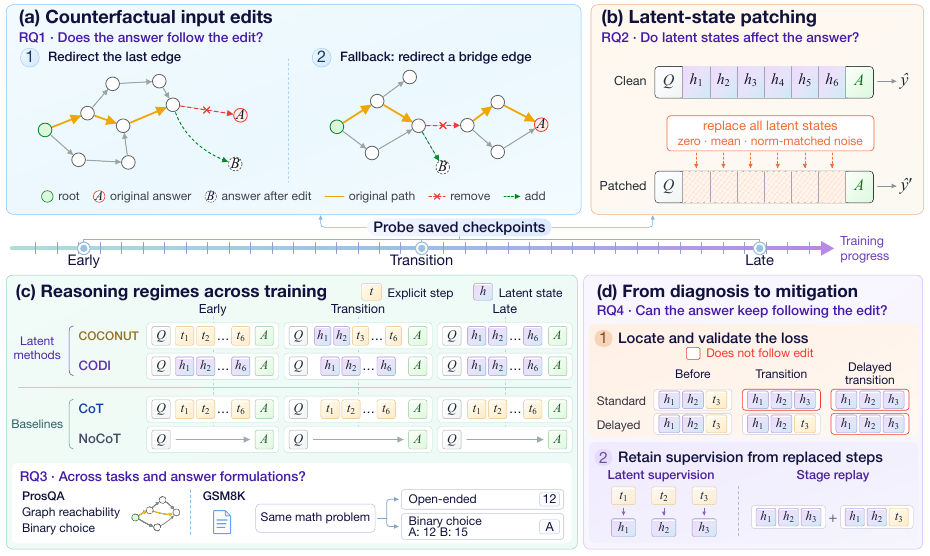}
\caption{%
    \textbf{Faithfulness probes across training.} (a) Verified graph edits change the ProsQA answer from $A$ to $B$. (b) All latent states are replaced with input and weights fixed. (c) Saved checkpoints of COCONUT, CODI and two baselines are probed; RQ3 adds two GSM8K answer formulations. (d) Runs branching at a COCONUT transition test whether the loss follows it and whether retaining supervision from replaced steps reduces it.}
\label{fig:teaser}
\end{figure*}

Chain-of-thought (CoT) reasoning represents intermediate computation as natural-language tokens, improving multi-step reasoning by exposing a step-by-step trace~\citep{wei2022_cot,nye2021_scratchpad}. Encoding each intermediate step as text comes at a cost: every reasoning step must be committed to discrete tokens, and the resulting traces lengthen inference. \emph{Latent reasoning} methods instead perform multi-step inference entirely in the model's continuous hidden states, promising more compact and efficient reasoning by removing the explicit trace that chain-of-thought requires~\citep{hao2025_COCONUT,shen2025_codi}.

This opacity raises a question of faithfulness analogous to that studied in explicit CoT, where generated rationales need not reflect the underlying computation~\citep{turpin2023_unfaithful,lanham2023_measuring_faithfulness}. Latent reasoning produces no such rationale to inspect, so the question is reframed as whether predictions follow the required reasoning path and how the latent states contribute to the final answer:\textit{ do interventions that change or erase these steps actually change the prediction?} Prior work reports that latent states can be weakly or unevenly used in the final prediction~\citep{zhang2025_do_latent_tokens_think,cui2026_supervision,li2026_dynamics}, and that models can rely on shortcuts instead of intermediate reasoning~\citep{lin2025_implicit_shortcut}.

These studies primarily characterize trained models or isolated training stages. They show that latent states can be weakly used, but not how this arises during training. An endpoint analysis therefore offers little guidance on which part of the training procedure is responsible or how to change it. Addressing these questions requires tracing behavioral and activation-level evidence throughout training. Such analysis is established for other model properties~\citep{tirumala2022_memorization_dynamics,tigges2024_circuits_across_training}, but has not yet been applied to latent reasoning faithfulness.

We therefore study whether strong task performance develops alongside faithful use of the required reasoning path, and how latent states contribute to answers during training. By following training, we aim to locate when gaps between task performance and faithful reasoning emerge and relate them to how latent reasoning is trained. We address four research questions using GPT-2 (124M) and Llama-1B models.

\noindent \textbf{RQ1: Do accuracy gains during training track improved responses to changes in the required reasoning path?}
High accuracy alone leaves open whether predictions use the reasoning-relevant information that determines the answer. On the graph reachability task ProsQA, breadth-first search (BFS) verifies that the edit in Figure~\ref{fig:teaser}(a) changes the correct answer by changing the required reasoning path. We measure whether an originally correct prediction follows this change. Applying this probe across training checkpoints, as in Figure~\ref{fig:teaser}(c), we find that accuracy can rise while counterfactual responsiveness falls, with distinct trajectories across training approaches.

\noindent \textbf{RQ2: Does the behavioral trajectory identified in RQ1 have an activation-level counterpart?}
To examine how latent reasoning states influence the prediction, we replace these states and measure the resulting output changes, as shown in Figure~\ref{fig:teaser}(b), following causal mediation and activation-patching analyses~\citep{vig2020_causal_mediation,heimersheim2024_patching}. On ProsQA with GPT-2, output sensitivity to noise replacement declines alongside counterfactual responsiveness.

\noindent \textbf{RQ3: How do activation-level trajectories compare across answer formulations?}
RQ1 and RQ2 identify declining sensitivity in the binary-choice ProsQA setting. As illustrated in Figure~\ref{fig:teaser}(c), we test whether this pattern extends to another task and its answer formulations by replacing all latent states in GPT-2 models trained separately on choice and open-ended versions of the same GSM8K math problems. The two formulations show opposite sensitivity trajectories, so the decline does not extend to both settings.

\noindent \textbf{RQ4: Can the trajectory diagnosis be used to reduce the loss of responsiveness?}
The trajectory analysis locates a loss of counterfactual responsiveness when reasoning becomes fully latent. This diagnosis motivates the two steps in Figure~\ref{fig:teaser}(d). First, using the standard schedule as a reference, we delay the transition to test whether the loss shifts with it, probing the transition's causal role. Delaying the loss serves as a diagnostic test rather than a remedy. Second, we test two ways to retain supervision from replaced steps: direct step-level supervision at latent positions, and replay of earlier-stage examples that still contain explicit reasoning steps. The loss shifts with the delayed transition, and step-level supervision raises responsiveness in later training stages.

\section{Related Work}
\label{sec:related}

\paragraph{Latent reasoning methods.}
Latent reasoning replaces some or all discrete chain-of-thought steps~\citep{wei2022_cot,nye2021_scratchpad} with computation in continuous hidden-state space. COCONUT~\citep{hao2025_COCONUT} feeds the model's last hidden state back as the next input embedding under a staged curriculum, while CODI~\citep{shen2025_codi} self-distils an explicit-CoT teacher into fixed-length continuous thoughts; related variants add non-verbal computation through pause tokens, recurrent depth, or implicit-CoT training~\citep{goyal2024_pause,geiping2025_recurrent,deng2024_implicit_cot,chen2025_survey_latentcot}. We study COCONUT and CODI as representatives of staged curricula and self-distillation, and follow each across training rather than at a single checkpoint.

\paragraph{Faithfulness for latent reasoning.}
Faithfulness diagnoses for explicit chain-of-thought assess whether the generated rationale reflects the model's underlying computation~\citep{jacovi-goldberg-2020-towards,turpin2023_unfaithful,lanham2023_measuring_faithfulness,lyu-etal-2024-towards}. For latent reasoning, the question becomes whether the continuous thought steps causally drive the final prediction. Studies of trained models or selected checkpoints suggest these continuous steps may be weakly used: ablation and perturbation on continuous thoughts largely preserve predictions~\citep{zhang2025_do_latent_tokens_think,dilgren2026_lrm_interp}, models exploit shortcuts over intermediate steps~\citep{lin2025_implicit_shortcut,liang2026_latentcot_step,cui2026_supervision}, and causal influence across latent positions is heterogeneous and can be non-local~\citep{li2026_dynamics}. None of them jointly tracks behavioral and activation-level evidence across training. We do so with perturbation tests developed for explicit reasoning~\citep{atanasova2023_counterfactual,lanham2023_measuring_faithfulness,paul-etal-2024-making} and causal-mediation analyses~\citep{vig2020_causal_mediation,heimersheim2024_patching}.

\paragraph{Training-time dynamics.}
Neural-network capabilities can change abruptly during training, as documented by work on grokking, induction-head phase transitions, and emergent abilities at scale~\citep{power2022_grokking,nanda2023_progress,olsson2022_induction_heads,wei2022_emergent}. \citet{tirumala2022_memorization_dynamics} track per-example behaviour across training rather than at convergence, and \citet{tigges2024_circuits_across_training} extend circuit analyses across checkpoints and scales. We apply this view to latent reasoning faithfulness and use the resulting trajectory to choose where to change training.

\section{Accuracy and Counterfactual Responsiveness Across Training}
\label{sec:rq1}

We track accuracy and counterfactual responsiveness across training approaches and scales.

\subsection{Experimental Setup}

\paragraph{Models and task.}
We compare four training paradigms on GPT-2 (124M), shown in Figure~\ref{fig:teaser}(c). CoT generates explicit reasoning, and answer-only (NoCoT) predicts the answer directly. COCONUT~\citep{hao2025_COCONUT} feeds hidden states back as input embeddings and replaces one more explicit step with a latent position every five epochs, up to six. Stage~$k$ denotes the epochs trained with $k$ latent positions. CODI~\citep{shen2025_codi} uses projected hidden states, a fixed six-position loop and answer-position self-distillation. Figure~\ref{fig:mechanisms} in Appendix~\ref{app:training} details both latent mechanisms. Both latent methods are also evaluated on Llama-3.2-1B-Instruct. All models are trained on ProsQA, which expresses directed graph reachability in natural language. ProsQA is chosen for three reasons. \emph{(a)} Its BFS solver verifies edited answers automatically, unlike manually labeled counterfactual sets~\citep{kaushik2020_counterfactual,gardner-etal-2020-evaluating}. \emph{(b)} The original COCONUT report uses ProsQA to argue for latent BFS, so edits probe that capability. \emph{(c)} Latent reasoning's reported advantage over explicit CoT is largest here, making low responsiveness a challenge to reasoning claims in a setting with strong task performance. Following prior work~\citep{cui2026_supervision}, we also evaluate CODI on ProsQA to compare its training trajectory with COCONUT on the same reasoning task. Appendix~\ref{app:training} lists the evaluated checkpoints and training settings.

\paragraph{Counterfactual edit on ProsQA.}
For each test instance with input $x_i$ and oracle answer $A_i$, let $B_i$ be the other candidate in the original two-way query. We construct a minimally edited input $x'_i$ whose oracle answer is $B_i$, and treat the model as counterfactually responsive on this pair when its prediction changes from $A_i$ to $B_i$ in response to the edit.
As illustrated in Figure~\ref{fig:teaser}(a), the construction first tries the gold path's last edge, then bridge edges whose removal disconnects $A_i$ from the query root $r_i$. BFS accepts a swap only if $A_i$ becomes unreachable and $B_i$ reachable. This yields $405$ pairs from $500$ test examples. The other $95$ have redundant paths that prevent a single-edge answer change. Every model and checkpoint uses the same $405$ pairs. Appendix~\ref{app:perturbation} gives the procedure as Algorithm~\ref{alg:perturbation}, together with the construction counts and reasoning-hop distribution.

\paragraph{Metrics.}
At each checkpoint, we label each original/edited prediction pair over $\{A,B,O\}^2$, where $A=A_i$, $B=B_i$, and $O$ denotes any other answer. $\mathrm{AB}$ is correct on both inputs; $\mathrm{AA}$ retains the original correct answer after editing. Contrast consistency (CC)~\citep{gardner-etal-2020-evaluating} measures joint correctness:
\[
\mathrm{CC} = \Pr\bigl(\hat{y}(x_i)=A_i \,\land\, \hat{y}(x'_i)=B_i\bigr),
\]
Counterfactual responsiveness (CFR) conditions on an originally correct prediction:
\[
\begin{aligned}
\mathrm{CFR}
&= \Pr\bigl(\hat{y}(x'_i)=B_i \mid \hat{y}(x_i)=A_i\bigr) \\
&= \mathrm{CC} / \mathrm{acc},
\end{aligned}
\]
where $\mathrm{acc} = \Pr(\hat{y}(x_i)=A_i)$. An originally correct prediction that faithfully follows the required reasoning path should follow the BFS-verified answer change, and CFR measures this necessary behavioral consequence. We also report unconditional edited accuracy, $\Pr[\hat y(x'_i)=B_i]$.

\subsection{Experimental Result}
\textbf{Validation-selected checkpoints show lower counterfactual responsiveness than CoT.}
At validation-selected checkpoints, COCONUT has the highest clean accuracy, but its $\mathrm{CFR}$ is $0.166$, versus CoT's $0.725$. CODI's $\mathrm{CFR}$ is $0.038$, closer to NoCoT's $0.003$. Both latent methods are less responsive than CoT, as Figure~\ref{fig:rq1-clean-vs-cfr} in Appendix~\ref{app:additional-results} shows. One possible explanation is weaker supervision of intermediate steps: COCONUT progressively removes textual targets, while CODI distils answer-position representations. Prior work links intermediate supervision to reduced shortcuts and reports shortcut reliance in COCONUT~\citep{cui2026_supervision,zhang2025_do_latent_tokens_think}.

\begin{figure}[t]
\centering
\includegraphics[width=\linewidth]{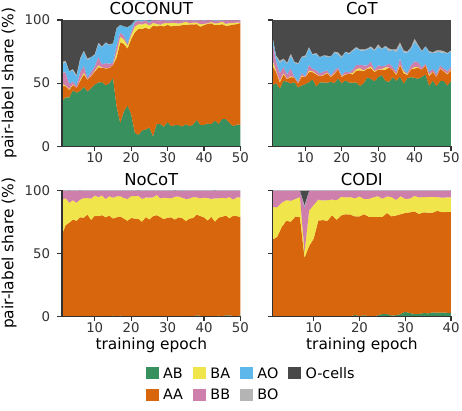}
\caption{Per-epoch joint label distribution on the $405$ ProsQA oracle pairs for the four training paradigms. Each column is a stacked area over $\{\mathrm{AB}, \mathrm{AA}, \mathrm{BA}, \mathrm{BB}, \mathrm{AO}, \mathrm{BO}, \mathrm{O}\textrm{-cells}\}$, where $\mathrm{O}\textrm{-cells}$ groups $\mathrm{OO}$, $\mathrm{OA}$, and $\mathrm{OB}$.}
\label{fig:rq2-coconut-pattern}
\end{figure}

\noindent \textbf{Clean accuracy can hide declining counterfactual responsiveness.}
The full COCONUT trajectory changes the interpretation (Figure~\ref{fig:rq2-coconut-pattern}). On GPT-2 small, clean accuracy rises from $72.8\%$ to $90.6\%$ across stages 2--3 (epochs~11--20), while CFR falls from $69.5\%$ to $28.3\%$. Unconditional edited accuracy also falls, from $60.7\%$ to $27.9\%$, confirming that performance on edited inputs also deteriorates. The joint labels move from AB toward AA, so the model increasingly retains its original answer after the verified graph edit. This change is not visible from the accuracy curve. Two further training seeds show the same fall at the transition into stage~3, as Figure~\ref{fig:seeds} in Appendix~\ref{app:additional-results} shows.

\noindent \textbf{CODI retains the original answer from early checkpoints.}
CODI follows a different GPT-2 trajectory (Figure~\ref{fig:rq2-coconut-pattern}): its $\mathrm{CFR}$ is zero at epoch~1 and remains at or below $0.0065$ through epoch~18; even the validation-selected epoch~37 reaches only $0.038$. Final checkpoints conceal this contrast: COCONUT loses earlier responsiveness, whereas CODI is weakly responsive from the start.

\begin{figure*}[t]
\centering
\includegraphics[width=\textwidth]{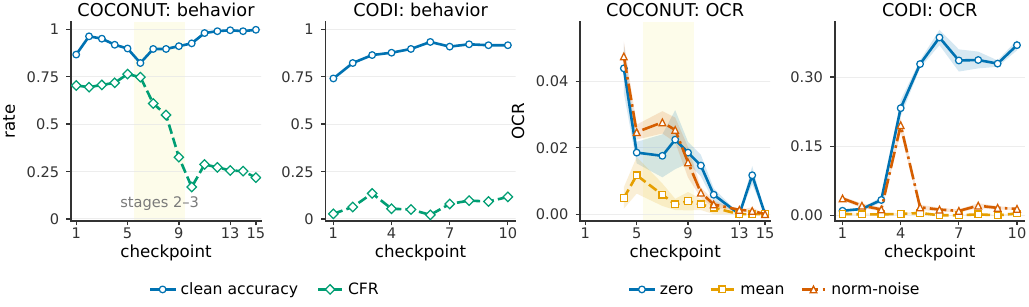}
\caption{Llama-1B ProsQA trajectories. The first two panels show clean accuracy and CFR; the last two show OCR under zero, mean, and norm-noise replacement. Shading marks COCONUT stages 2--3. OCR bands show one standard deviation across four evaluation subsamples; the two OCR panels use different vertical scales.}
\label{fig:llama1b-scale-replication}
\end{figure*}

\noindent \textbf{The timing contrast persists with Llama-1B.}
Figure~\ref{fig:llama1b-scale-replication} reproduces this contrast with Llama-3.2-1B-Instruct. During COCONUT stages 2--3, accuracy rises while CFR falls by $42.3$ percentage points. CFR continues to decline as more explicit steps are replaced. CODI remains weakly responsive as accuracy improves. Across scales, this contrast aligns with how latent training is introduced: COCONUT adds recurrent states stage by stage, whereas CODI applies a fixed six-state loop and answer-position self-distillation from the outset. This alignment points to the training schedule as a possible explanation for when the gap emerges, alongside differences in latent feedback and objectives.

\paragraph{Control for general failure under edits.}
We compare premise edits that preserve or change the answer on the same $345$ examples. At epoch~48, accuracy is $96.81\%$ for preserving edits and $20.87\%$ for changing edits. Models therefore remain accurate under some premise edits. Appendix~\ref{app:controls} gives the comparison and graph validity checks.

\section{Latent-State Sensitivity Across Training}
\label{sec:rq2}

RQ1 reveals weaker responses to changes in the required reasoning path. We test whether latent states also have less influence on answers, first across checkpoints and then within groups of examples defined by their behavior.

\subsection{Experimental Setup}

\paragraph{Patching across checkpoints.}
Following causal mediation and activation patching analyses~\citep{vig2020_causal_mediation,heimersheim2024_patching}, at each checkpoint we replace all latent reasoning states with \texttt{zero}, the corresponding-position mean (\texttt{mean}), or norm-matched Gaussian noise (\texttt{norm-noise}), as in Figure~\ref{fig:teaser}(b). Downstream states and the answer are then generated under the intervention; details are in Appendix~\ref{app:metrics}.

The first metric is the output change rate, which measures the probability that predictions change under intervention:
\begin{equation}
\mathrm{OCR} = \Pr\bigl[\hat{y}^{\text{patch}}(x) \neq \hat{y}^{\text{clean}}(x)\bigr].
\end{equation}
The second metric is the preserved-when-correct rate, which isolates the subset of inputs that are correct in the clean pass to measure how often the intervention is destructive:
\begin{equation}
\mathrm{PWC} = \Pr\bigl[\hat{y}^{\text{patch}}(x) = y \,\bigm|\, \hat{y}^{\text{clean}}(x) = y\bigr].
\end{equation}
$\mathrm{OCR}$ counts every change, including changes to already-incorrect answers, whereas $\mathrm{PWC}$ isolates damage to correct ones. We report the mean and standard deviation across four evaluation subsamples. Further details are in Appendix~\ref{app:metrics}.

\paragraph{Patching by cohort.}
The aggregate movement from $\mathrm{AB}$ to $\mathrm{AA}$ in RQ1 leaves open whether the behavioral and activation changes involve the same examples. We define three cohorts from epochs~15, 16 and~48: stable pairs are $\mathrm{AB}$ at epochs~15 and~48, drift pairs move from $\mathrm{AB}$ at epoch~15 to $\mathrm{AA}$ at epochs~16 and~48, and retention pairs are $\mathrm{AA}$ at epochs~15 and~48. For each cohort, we measure how much norm-noise replacement lowers the binary log-probability margin, the log probability of the unpatched predicted candidate minus that of the other candidate, with the generated answer prefix held fixed within each checkpoint. Positive values mean that patching weakens the prediction, including when the greedy answer is unchanged. Appendix~\ref{app:metrics} gives the details.

\subsection{Experimental Result}
\begin{figure}[t]
\centering
\includegraphics[width=\linewidth]{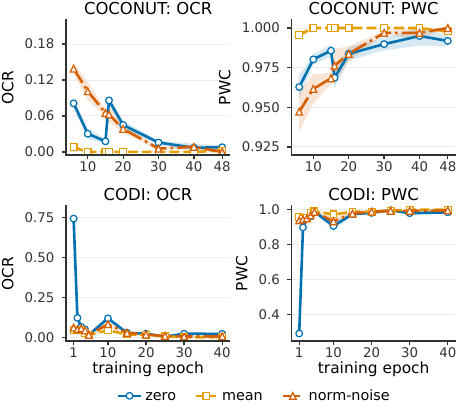}
\caption{Output sensitivity to replacing latent states along the ProsQA training trajectory. Top: COCONUT under \texttt{zero}, \texttt{mean}, and \texttt{norm-noise}. Bottom: CODI on its 11-checkpoint grid. Lines show means across four 256-question evaluation subsamples. Bands show one standard deviation.}
\label{fig:rq3-prosqa-patching}
\end{figure}

\textbf{Declining counterfactual responsiveness has an activation-level counterpart on ProsQA.}
Figure~\ref{fig:rq3-prosqa-patching} reports $\mathrm{OCR}$ and $\mathrm{PWC}$ for the GPT-2 comparison. For both COCONUT and CODI, all three replacements approach low OCR and high PWC by late checkpoints. This result provides an activation-level counterpart to RQ1 on the shared backbone. The Llama-1B replication below shows that the conclusion is replacement-specific at larger scale.

\noindent \textbf{The two procedures reach low sensitivity at different points during training.}
For COCONUT, $\mathrm{OCR}$ falls and $\mathrm{PWC}$ rises across the curriculum. CODI reaches low sensitivity earlier, aside from a brief epoch-$10$ transient. This timing matches the behavioral difference in RQ1. COCONUT loses counterfactual responsiveness during training, while CODI starts with low responsiveness and reaches low activation sensitivity early.

\noindent \textbf{The Llama-1B activation trajectories preserve the method difference.}
The two right panels of Figure~\ref{fig:llama1b-scale-replication} show the Llama-1B results. For COCONUT, norm-noise $\mathrm{OCR}$ falls from $4.75\%$ at the first evaluated latent checkpoint to $0\%$ at the final checkpoint. Across the ten checkpoints shared by the behavioral and activation analyses, CFR and norm-noise OCR have a Pearson correlation of $0.888$. CODI has low final OCR under mean and norm-noise replacement, at $0.39\%$ and $1.40\%$, respectively. Zero replacement instead reaches $36.91\%$. This shows that output sensitivity depends on the scale and content of the replacement.

\noindent \textbf{Examples that stop following the edit also become less sensitive to norm-noise replacement.}
Table~\ref{tab:rq2-cohort} reports the median margin decrease for each cohort. On $G_{\text{drift}}$ under all-step \texttt{norm-noise}, the median decrease on the perturbed input falls from $1.71$ to $0.58$ between epochs~15 and 16, reaching $0.20$ at epoch~48. The drift cohort follows a different pattern on original inputs across this boundary. Sensitivity also decreases in the stable cohort. Appendix~\ref{app:controls} reports paired uncertainty estimates.

\begin{table}[h]
\centering
\small
\begin{tabular}{llccc}
\toprule
Cohort & Input & ep 15 & ep 16 & ep 48 \\
\midrule
$G_{\text{stable}}$ ($n=41$)  & perturbed & 1.93 & 0.92 & 0.13 \\
$G_{\text{drift}}$  ($n=68$)  & original  & 1.87 & 2.38 & 0.20 \\
$G_{\text{drift}}$  ($n=68$)  & perturbed & \textbf{1.71} & \textbf{0.58} & 0.20 \\
$G_{\text{retain}}$ ($n=37$)  & perturbed & 2.31 & 3.37 & 0.25 \\
\bottomrule
\end{tabular}
\caption{Median margin decrease under all-step \texttt{norm-noise}, with the baseline answer prefix fixed. Each pair averages five trials. Positive values mean patching weakens the baseline prediction. Sizes are nominal; finite-margin counts are in Appendix~\ref{app:controls}.}
\label{tab:rq2-cohort}
\end{table}

\paragraph{Intervention controls.}
The later OCR decline persists when norm-noise is restricted to position~0 or the first two positions at epochs~15, 16 and~48. Transferring states between original and edited inputs rarely changes answers (Appendix~\ref{app:controls}).

\section{Latent-State Sensitivity Across Answer Formats}
\label{sec:rq3}

ProsQA presents two answer candidates. We next ask whether its declining sensitivity extends to another task and its answer formulations. Prior work reports that latent reasoning steps show different levels of involvement across datasets~\citep{li2026_dynamics,cui2026_supervision}. We compare binary-choice and open-ended formulations of the same GSM8K problems across training.

\begin{figure*}[t]
\centering
\includegraphics[width=\textwidth]{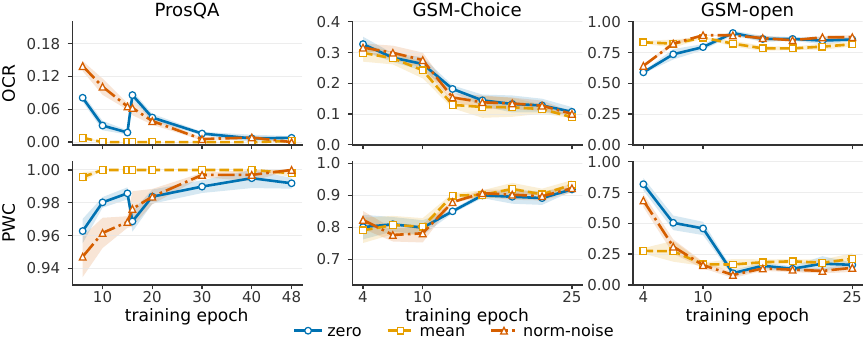}
\caption{Output sensitivity under three interventions that replace all latent states in COCONUT. Left: ProsQA. Middle: GSM-Choice. Right: GSM-open. Top: output change rate $\mathrm{OCR}$. Bottom: preserved-when-correct rate $\mathrm{PWC}$. Lines show means across four evaluation subsamples. Bands show one standard deviation.}
\label{fig:rq4-three-tasks}
\end{figure*}

\subsection{Experimental Setup}
We apply all-step activation patching to separately trained COCONUT models on two answer formulations of the same GSM8K problems~\citep{cobbe2021_gsm8k}. \emph{GSM-Choice} adopts a binary-choice formulation analogous to ProsQA, presenting the gold answer and a distractor and predicting the answer letter, whereas \emph{GSM-open} requires open-ended numerical generation. Both use the same backbone, optimizer, hyperparameters, curriculum and decoding. Each follows the standard COCONUT procedure~\citep{hao2025_COCONUT} on its own formulation, in which a CoT model is first trained on the same data and its checkpoint with the best validation accuracy initializes COCONUT training. Details are in Appendices~\ref{app:training} and~\ref{app:metrics}.

\subsection{Experimental Result}
\textbf{The two GSM formulations show opposite trajectory directions.}
Figure~\ref{fig:rq4-three-tasks} shows the main result. On GSM-open, $\mathrm{OCR}$ under \texttt{norm-noise} rises during training and remains high at the final evaluated epoch. On GSM-Choice, the same intervention moves in the opposite direction: $\mathrm{OCR}$ falls and $\mathrm{PWC}$ rises across training, matching the direction observed on ProsQA in \S\ref{sec:rq2}. As a control with replacements drawn from the model itself, we transfer latent states between questions. Replacing all latent states preserves the opposite endpoint trends, as Appendix~\ref{app:controls} reports.

\noindent \textbf{The model ranking also differs between the two formulations.}
At validation-selected checkpoints, COCONUT outperforms CoT on GSM-Choice, where intervention sensitivity declines, but underperforms CoT on GSM-open, where sensitivity rises. An accuracy advantage over CoT can therefore coexist with declining intervention sensitivity. On GSM-Choice, \texttt{zero} replacement also corrects $18.0\%$ of initially wrong predictions. Accuracy and the influence of latent states provide different views of model behavior. Table~\ref{tab:rq4-gsm-format-ranking} in Appendix~\ref{app:additional-results} reports the accuracies.

\section{From Diagnosis to Mitigation}
\label{sec:rq4}

RQ1 locates the loss of counterfactual responsiveness at COCONUT's stage transitions. RQ2 shows that the influence of latent states on the answer declines over the same period and in the same examples, and RQ3 shows that this decline does not occur in every setting. Guided by this diagnosis, we modify training to reduce the loss.

\subsection{Experimental Setup}
We test one change that probes the cause of the loss and two that aim to reduce it.

\paragraph{Delayed transition.}
RQ1 observes the loss at a stage transition but cannot separate the transition from continued training. This run holds stage~2 for six more epochs before entering stage~3. If the transition is the cause, the loss moves with it.

\paragraph{Latent supervision.}
At a transition, the text target of the replaced step is removed, and the latent state that takes its place is trained only through later tokens. This run restores a target at that position. The state fed into latent position~$j$ is trained, through the language-model head, to predict the first token of the concept reached by replaced step~$j$.

\paragraph{Stage replay.}
Before a transition, the model follows the edit while the step is explicit. This run mixes earlier-stage examples into training, so that a step is seen both as text and as a latent state. At stage~$k$, each example is trained with probability $0.3$ at an earlier stage that keeps a step explicit.

\paragraph{Runs and evaluation.}
All runs keep the training settings of the GPT-2 COCONUT run in \S\ref{sec:rq1} and branch from its epoch-15 checkpoint, the last of stage~2. Apart from the delayed run, they follow a compressed curriculum with two epochs for each of stages~3--6, and a \emph{standard} run with no other change serves as the reference. A second set of runs branches from epoch~5, before any step is replaced, so that each change applies from the first latent stage. It contains a standard run, latent supervision, stage replay, and a \emph{combined} run that applies both changes. Each checkpoint is evaluated on the $405$ pairs in two ways. \emph{Fully latent} inference uses as many latent positions as the training stage. \emph{Partly explicit} inference uses two latent positions, so that later steps are generated as text. Appendix~\ref{app:rq4} defines both objectives formally and lists the schedules.

\begin{figure}[t]
\centering
\includegraphics[width=\linewidth]{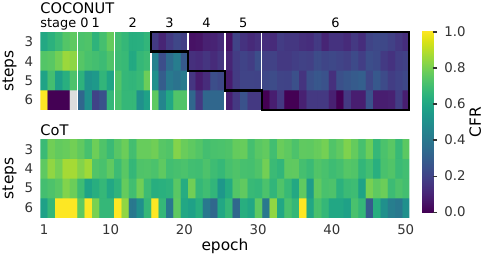}
\caption{CFR by number of gold reasoning steps at every epoch, for the original GPT-2 COCONUT run and for CoT. The black outline encloses the epochs at which a chain of that length is fully latent. The four rows contain $135$, $191$, $67$ and $12$ pairs.}
\label{fig:rq4-steps}
\end{figure}

\subsection{Experimental Result}
\textbf{The loss tracks the replacement of explicit steps.}
Stage~$k$ replaces the first $k$ reasoning steps, so shorter chains become fully latent earlier. Figure~\ref{fig:rq4-steps} shows CFR for each chain length at every epoch. In stage~3, CFR falls most for three-step examples, whose reasoning is now fully latent, and less for the longer chains that retain explicit steps. CoT shows no such separation.

\begin{figure}[t]
\centering
\includegraphics[width=\linewidth]{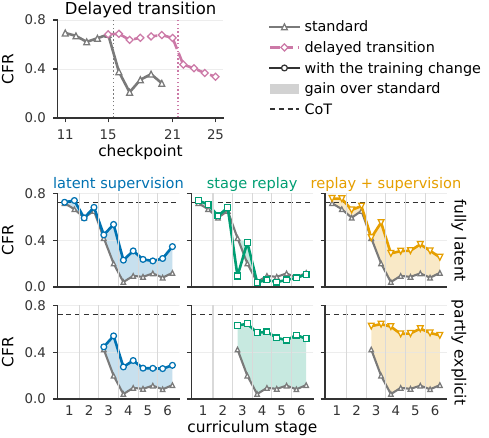}
\caption{Diagnosis-guided runs of GPT-2 COCONUT on ProsQA. Top: CFR under the standard and delayed transitions, marked by dotted lines. Below: CFR at every checkpoint of the runs that branch from epoch~5. Each column compares one training change with the standard run, and the shaded area is its gain.}
\label{fig:rq4-intervention}
\end{figure}

\noindent \textbf{Delaying the transition delays the loss.}
The top panel of Figure~\ref{fig:rq4-intervention} shows that CFR stays at its stage-2 level across the six added epochs, although the optimizer is still recreated every epoch. It then falls in the first stage-3 epoch, as in the standard run six epochs earlier. The loss thus follows the transition.

\noindent \textbf{Latent supervision raises responsiveness at the end of the curriculum.}
The first column of the grid in Figure~\ref{fig:rq4-intervention} compares latent supervision with the standard run. Under fully latent inference, it does not prevent the loss, but from stage~4 onward its CFR stays above the standard run. Its final CFR is $0.30$ against $0.10$ for the standard run, with higher accuracy, as in the branch from epoch~15.

\noindent \textbf{Replay preserves responsiveness when a step stays explicit.}
The second column shows stage replay. It matches the standard run under fully latent inference. Under partly explicit inference, its CFR is more than four times that of the standard run, at lower accuracy, and the branch from epoch~15 gives the same result. Replay thus retains this ability, which is unused once the chain is fully latent.

\noindent \textbf{The combined run gains in both accuracy and responsiveness.}
The third column shows the combined run. Under fully latent inference, its accuracy is $0.96$, above the standard run at $0.90$ and CoT at $0.76$, and its CFR is $0.28$ against $0.10$ for the standard run. Under partly explicit inference, its CFR reaches $0.56$ at an accuracy of $0.88$. CoT still has the highest CFR, the dashed line in Figure~\ref{fig:rq4-intervention}. Table~\ref{tab:rq4-summary} in Appendix~\ref{app:rq4} lists all values.

\noindent \textbf{What the trajectory contributes.}
A final checkpoint shows that responsiveness is low, which is compatible with many remedies. The trajectory adds three things. It shows that responsiveness is present before the transition and lost at it, so the remedy is to preserve it. It identifies what is removed at that point, the text target of the replaced step, which fixes where the supervision is placed and what it predicts. It also provides the checkpoints from which the delayed and branched runs start, and the earlier-stage format for replay.

\section{Conclusion}

Tracing latent reasoning faithfulness across training shows when accuracy and responsiveness diverge. In GPT-2 (124M) and Llama-1B models, COCONUT loses counterfactual responsiveness as accuracy improves, whereas CODI is weakly responsive from its earliest checkpoints, which final checkpoints do not show. The trajectory also indicates where to change training. COCONUT's loss follows the curriculum transitions that replace explicit steps, and supervising the replacing latent states more than doubles final counterfactual responsiveness without lowering accuracy.

\newpage
\section*{Limitations}
\label{sec:limitations}

The verified counterfactual analysis applies to tasks with a deterministic oracle for the perturbed input. ProsQA provides this oracle through a BFS solver. Other tasks require suitable counterfactual constructions. Valid edited graphs may still differ from the training distribution and affect generalization. CFR measures whether a model that answers correctly on the original input follows the answer change after editing. High CFR may also reflect shortcuts. Latent interventions provide complementary evidence about the influence of hidden states. Identifying their reasoning content requires further analysis.

Resource constraints limit our experiments to GPT-2 (124M) and Llama-1B, trained separately on two tasks. We therefore prioritize additional training seeds for GPT-2 COCONUT on ProsQA, the setting on which the diagnosis and the training changes in \S\ref{sec:rq4} are based. The remaining trajectories use one training run per condition, with uncertainty estimated through evaluation sampling. The training changes in \S\ref{sec:rq4} are tested on one task and one model, with runs from two branch points, and serve as a proof of concept. COCONUT and CODI use different training settings, and our CODI adaptation to ProsQA follows prior work without validation on CODI's published benchmarks. We interpret differences between their trajectories in the context of these settings. COCONUT also recreates its optimizer at every epoch; CFR changes little between epochs within a stage and falls at stage transitions (Appendix~\ref{app:additional-results}), but resets may still interact with the curriculum. The GSM comparison is descriptive. Separately trained models differ in answer formulation and distractors, which limits causal attribution to answer format. Each is initialized from a CoT model trained on its own formulation, as in the COCONUT reference implementation. Future work can test generality across seeds, tasks, mixed training and larger models.

\bibliography{custom}

\appendix   
\section{Appendix}

\subsection{Training and Checkpoint Details}\label{app:training}

Each GPT-2 training run uses one GPU. The COCONUT, CoT, and NoCoT runs use an NVIDIA RTX PRO 6000 Blackwell Server Edition, while the CODI run uses an NVIDIA GeForce RTX 5090. The Llama-1B COCONUT and CODI runs use four and one 48GB NVIDIA GeForce RTX 4090 GPUs, respectively.

\begin{figure}[t]
\centering
\includegraphics[width=\linewidth]{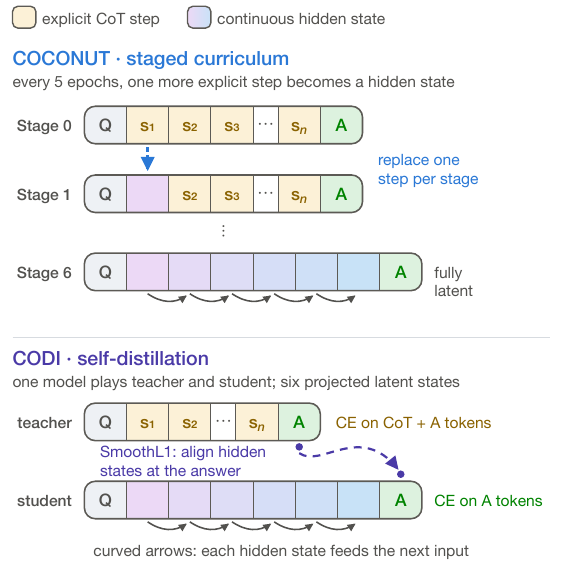}
\caption{%
\textbf{Training mechanisms of the two latent reasoning paradigms.}
COCONUT replaces one explicit step with a continuous hidden state every five epochs until the chain is fully latent.
CODI uses one model as both teacher and student in a fixed six-state loop, with chain/answer cross entropy and answer-position SmoothL1 state alignment.}
\label{fig:mechanisms}
\end{figure}

\paragraph{Implementation.}
CoT, NoCoT, and COCONUT use the COCONUT reference training loop~\citep{hao2025_COCONUT}. We change local paths, use gradient accumulation to obtain an effective batch of $128$, retain every checkpoint, and apply a compatibility change to the key-value cache representation. The AdamW hyperparameters remain unchanged. The CODI reference repository~\citep{shen2025_codi} does not include ProsQA. We add local ProsQA loading, validation, and answer formatting, together with compatibility changes for the Trainer interface; the CODI training objective is unchanged. All reported models use training seed~$0$, except the two additional seeds in Appendix~\ref{app:additional-results}.

\paragraph{Compute and artifacts.}
The training budgets are approximately $36$ GPU-hours for GPT-2 ProsQA $\times$ COCONUT, $83$ GPU-hours each for GSM-open $\times$ COCONUT and GSM-Choice $\times$ COCONUT, $5$ GPU-hours for GPT-2 ProsQA $\times$ CODI, $38$ GPU-hours for Llama-1B COCONUT, and $6$ GPU-hours for Llama-1B CODI. The additional GPT-2 ProsQA runs take approximately $5$ GPU-hours for the delayed transition, $7$ GPU-hours for each run that branches from epoch~15, $8$ GPU-hours for each run that branches from epoch~5, and $12$ GPU-hours for each additional training seed through epoch~25. The GPT-2 checkpoint is MIT-licensed, and the other external artifacts are used under their respective licenses. We use these artifacts only for research training and evaluation, and the ProsQA perturbation pairs and GSM-Choice reformulation introduced here are intended as research evaluation artifacts. Checkpoint saving is not filtered by validation score. When an analysis uses a validation-selected checkpoint, we select it afterward and state that choice explicitly.

\paragraph{Backbone and tokeniser.}
The GPT-2 comparison uses \texttt{openai-community/gpt2}, which has 124M parameters~\citep{radford2019_gpt2}. The larger replications use \texttt{meta-llama/Llama-3.2-1B-Instruct} from the Llama~3 family~\citep{grattafiori2024_llama3}. Each model uses its native tokeniser. COCONUT adds \texttt{<|start-latent|>}, \texttt{<|end-latent|>}, and \texttt{<|latent|>}. CODI reserves three additional embedding indices for padding and the beginning and end of its latent loop.

\paragraph{Optimiser.}
For the GPT-2 CoT, NoCoT, and COCONUT runs, we use AdamW with learning rate $10^{-4}$ and weight decay $0.01$. Precision is FP32. COCONUT recreates AdamW at every epoch through \texttt{reset\_optimizer=True}. CoT and NoCoT keep the optimiser state. Llama-1B COCONUT uses BF16, learning rate $5 \cdot 10^{-5}$, and weight decay $0.1$. GPT-2 CODI uses AdamW with learning rate $3 \cdot 10^{-3}$, weight decay $0.1$, a cosine schedule, a $0.03$ warm-up ratio, gradient clipping at $2.0$, BF16 mixed precision, and LoRA adapters with $r=128$, $\alpha=32$, and dropout $0.1$ on frozen weights. Llama-1B CODI keeps the same LoRA and scheduler settings but uses learning rate $8 \cdot 10^{-4}$ and a distillation-loss multiplier of $20$. The CODI projection head has dimension $768$ for GPT-2 and $2048$ for Llama-1B, with dropout $0$. With \texttt{distill\_loss\_div\_std=True}, each layer's distillation loss is divided by the standard deviation of the corresponding reference answer-position hidden state.

\paragraph{ProsQA training.}
Train/valid/test splits contain $17{,}886$ / $300$ / $500$ examples respectively. The GPT-2 CoT, NoCoT, and COCONUT models are each trained for $50$ epochs on the same loader. The COCONUT curriculum uses $c_{\text{thought}}=1$, $\texttt{epochs\_per\_stage}=5$, and $\texttt{max\_latent\_stage}=6$. The latent token budget is $0$ for checkpoints~1--5, increases by one every five checkpoints, reaches $6$ at checkpoint~31, and remains at $6$ thereafter. CoT keeps the same scaffolding but with $\texttt{cot=True}$ (no latent positions). NoCoT uses $\texttt{no\_cot=True}$, $c_{\text{thought}}=0$, and $\texttt{max\_latent\_stage}=0$. CODI is trained for $40$ epochs with $\texttt{num\_latent}=6$, $\texttt{model\_max\_length}=512$, $\texttt{max\_token\_num}=1000$, $\texttt{include\_last\_cot=True}$, and \texttt{remove\_eos=True}, using per-device batch size $16$ with $8$ accumulation steps (effective batch $128$).

\paragraph{Llama-1B ProsQA replications.}
Both runs use the same ProsQA splits. COCONUT stage 0 trains explicit CoT for three epochs. It then trains stages 1--6 for two epochs each, yielding checkpoints 1--15. Each of four GPUs uses a batch size of $16$ with two gradient accumulation steps, giving an effective batch size of $128$. CODI trains for ten epochs with six latent positions throughout. It uses a per-device batch size of $8$ with $16$ accumulation steps on one GPU, also giving an effective batch size of $128$. We save every checkpoint in both runs.

\paragraph{GSM-open training.}
Train/valid/test splits contain $385{,}620$ / $500$ / $1{,}319$ examples (the \texttt{coconut/data/gsm\_train.json} augmentation expansion of GSM8K~\citep{cobbe2021_gsm8k}). Training has two phases. In the CoT phase, GPT-2 is trained for $25$ epochs with $c_{\text{thought}}=0$ and $\texttt{max\_latent\_stage}=0$, saving every epoch. We use \texttt{checkpoint\_19}, the highest-validation checkpoint of this run, to initialise COCONUT. Initializing COCONUT from a CoT checkpoint follows the reference implementation~\citep{hao2025_COCONUT}. The COCONUT phase uses $c_{\text{thought}}=2$, $\texttt{epochs\_per\_stage}=3$, $\texttt{max\_latent\_stage}=3$, and \texttt{resume=3}. Consequently, its training loop begins at checkpoint~$4$ and ends at checkpoint~$25$, for $22$ training epochs. The saved-checkpoint schedule is epochs~$4$--$6$ with two latent positions, epochs~$7$--$9$ with four, and epochs~$10$--$25$ with six. There are no COCONUT checkpoints for epochs~$1$--$3$. Both phases use \texttt{batch\_size\_training=32} and $\texttt{gradient\_accumulation\_steps}=4$, giving an effective batch of $128$.

\paragraph{GSM-Choice training.}
The GSM-Choice corpus uses the same $385{,}620$ / $500$ / $1{,}319$ split sizes, but each item is rewritten as a two-option multiple choice question following the methodology of \citet{zhang2024_gsm_mc}. The question text is appended with \texttt{$\backslash$nA) v$_1\backslash$nB) v$_2\backslash$nAnswer:}, where one of $\{v_1, v_2\}$ is the gold numerical answer and the other is a distractor. Distractors are drawn, in order of preference, from (i) the public GSM-MC option pool of \citet{zhang2024_gsm_mc} when available, (ii) intermediate values that appear inside the \texttt{<<\dots=N>>} step calculations, and (iii) magnitude-perturbed variants of the gold answer ($\pm 1, \pm 10, \times 2, /2$). The gold letter is balanced across positions ($\Pr[\text{gold}=A]=0.5$). Training uses the same two-phase protocol and hyperparameters as GSM-open. The CoT phase runs for $25$ epochs. The COCONUT phase starts from \texttt{cot-gsm-mc/checkpoint\_12}, the highest-validation checkpoint of the CoT run, and saves checkpoints~4--25.

\paragraph{Checkpoint grids for trajectory analyses.}
Behavioural analyses consume every saved checkpoint of each run: epochs~$1$--$50$ for GPT-2 CoT, NoCoT and COCONUT on ProsQA, epochs~$1$--$15$ for Llama-1B COCONUT, epochs~$1$--$10$ for Llama-1B CODI, epochs~$4$--$25$ for COCONUT on GSM-open and GSM-Choice, and epochs~$1$--$40$ for GPT-2 CODI on ProsQA. For the activation patching analyses we evaluate on a per-setting grid of checkpoints chosen to span training, with extra density placed where the trajectory in \S\ref{sec:rq2} changes most rapidly:
\begin{itemize}
\item \textbf{ProsQA $\times$ COCONUT}: epochs $\{6, 10, 15, 16, 20, 30, 40, 48\}$ ($8$ checkpoints with dense coverage of the middle curriculum stages);
\item \textbf{ProsQA $\times$ COCONUT with Llama-1B}: epochs $\{4, 5, 7, 8, 9, 10, 11, 13, 14, 15\}$ ($10$ checkpoints spanning all six latent stages);
\item \textbf{ProsQA $\times$ CODI}: epochs $\{1, 2, 3, 4, 5, 10, 15, 20, 25, 30, 40\}$ ($11$ checkpoints; the first five epochs capture the early drop in the $\mathrm{OCR}$ trajectory);
\item \textbf{ProsQA $\times$ CODI with Llama-1B}: epochs $\{1,\ldots,10\}$ (all saved checkpoints);
\item \textbf{GSM-open $\times$ COCONUT}: epochs $\{4, 7, 10, 13, 16, 19, 22, 25\}$ ($8$ checkpoints; the first saved checkpoint at each of two, four, and six latent positions, followed by five more checkpoints with six positions);
\item \textbf{GSM-Choice $\times$ COCONUT}: same epoch grid as GSM-open ($8$ checkpoints).
\end{itemize}
The activation patching sampling protocol, decoding settings, and uncertainty methodology are in \S\ref{app:metrics}.

\subsection{Counterfactual Edit Construction on ProsQA}\label{app:perturbation}

Algorithm~\ref{alg:perturbation} gives the construction procedure of \S\ref{sec:rq1}. It uses $(root,target,t^-)=(r_i,A_i,B_i)$.

\begin{algorithm}[t]
\caption{ProsQA perturbation verified by the BFS oracle.}
\label{alg:perturbation}
\begin{algorithmic}[1]
\Require example $(\mathcal{G}, root, target, t^-, \pi)$ with gold path $\pi$
\Ensure swap $(u, v) \to (u, t^-)$ that flips the oracle answer, or $\bot$
\State \textbf{Tier 1 (canonical last hop):}
\State \quad $(u, v) \gets (\pi_{|\pi|-2}, \pi_{|\pi|-1})$
\State \quad $\mathcal{G}' \gets (\mathcal{G} \setminus \{(u, v)\}) \cup \{(u, t^-)\}$
\If{$\neg \mathrm{BFS}(\mathcal{G}', root, target)$ \textbf{and} $\mathrm{BFS}(\mathcal{G}', root, t^-)$}
    \State \Return $(u, v, \texttt{canonical})$
\EndIf
\State \textbf{Tier 2 (bridge search):}
\State \quad $\mathcal{B} \gets \mathrm{Bridges}(\mathcal{G}, root, target)$
\For{$(u, v) \in \mathcal{B}$}
    \State $\mathcal{G}' \gets (\mathcal{G} \setminus \{(u, v)\}) \cup \{(u, t^-)\}$
    \If{$\neg \mathrm{BFS}(\mathcal{G}', root, target)$ \textbf{and} $\mathrm{BFS}(\mathcal{G}', root, t^-)$}
        \State \Return $(u, v, \texttt{bridge})$
    \EndIf
\EndFor
\State \Return $\bot$
\end{algorithmic}
\end{algorithm}

\paragraph{Solver primitives.}
Reachability is decided by breadth-first search on the directed adjacency list. An edge $(u, v)$ is a \emph{bridge for the (root, target) query} if removing it disconnects \textit{target} from \textit{root} (this is a path-specific notion, not the classical undirected bridge). Bridges are enumerated by trying each edge $(u, v)$ that lies on at least one \textit{root}~$\to$~\textit{target} path and re-running BFS without it.

\paragraph{Swap acceptance criteria.}
In addition to the answer change confirmed by the BFS oracle in Algorithm~\ref{alg:perturbation}, a candidate swap must preserve a valid directed graph and admit a single-sentence surface rewrite of the form ``Every $u$ is a $v$'' $\to$ ``Every $u$ is a $t^-$'', where $t^-$ is the alternative candidate $B_i$. Isolated nodes are not an exclusion criterion. The reachability checks and surface rewrite determine acceptance.

\paragraph{Strategy-level coverage.}
The $405$ accepted pairs split into $335$ found by the canonical last-hop swap ($82.7\%$ of accepted) and $70$ found by the bridge search ($17.3\%$). The rejected examples are those for which every edge on every $root \to target$ path has a bypass, so no single-edge swap can achieve the conjunction $\neg[root \to target] \land [root \to t^-]$. Because the original ProsQA graph satisfies $\neg[root \to t^-]$ by construction (only $target$ is reachable from $root$), the binding constraint is the existence of a bridge edge $(u, v)$ whose removal disconnects $target$; these rejected examples would require coordinated multi-edge edits, which we exclude in order to keep the surface change to a single rewritten sentence. Selection depends only on graph structure and is independent of model predictions.

\paragraph{Reasoning-hop distribution.}
Table~\ref{tab:hop-distribution} compares gold reasoning-step counts in the full test set and the accepted pairs. Acceptance is higher for longer paths, so the accepted set is enriched in $4$--$6$-step problems while retaining the full hop range.

\begin{table}[t]
\centering
\small
\setlength{\tabcolsep}{4pt}
\begin{tabular*}{\linewidth}{@{\extracolsep{\fill}}rccccc@{}}
\toprule
 & \multicolumn{2}{c}{Test} & \multicolumn{2}{c}{Pert.} & \multicolumn{1}{c}{Accept.} \\
\cmidrule(lr){2-3}\cmidrule(lr){4-5}
Steps & $n$ & \% & $n$ & \% & rate \\
\midrule
3 & 202 & 40.4 & 135 & 33.3 & 66.8\% \\
4 & 217 & 43.4 & 191 & 47.2 & 88.0\% \\
5 &  69 & 13.8 &  67 & 16.5 & 97.1\% \\
6 &  12 & \phantom{0}2.4 &  12 & \phantom{0}3.0 & 100\phantom{.0}\% \\
\midrule
total & 500 & & 405 & & 81.0\% \\
\bottomrule
\end{tabular*}
\caption{Distribution of the number of gold reasoning-step sentences (a proxy for canonical-path length) on the ProsQA test set ($n=500$) and on the $405$ accepted perturbation pairs. The accepted pairs span the full hop range, with greater representation of longer-path examples.}
\label{tab:hop-distribution}
\end{table}

\paragraph{Question rewriting.}
After accepting a swap $(u, v) \to (u, t^-)$, the perturbed natural-language question is generated by replacing the unique sentence ``Every $\sigma(u)$ is a $\sigma(v)$.'' with ``Every $\sigma(u)$ is a $\sigma(t^-)$.'' (where $\sigma$ is the symbol-name map of the example), leaving every other sentence unchanged. The query sentence (``Is $\sigma(root)$ a $\sigma(target)$ or a $\sigma(t^-)$?'') is preserved verbatim, so the answer flips from $\sigma(target)$ to $\sigma(t^-)$ but the candidate set, entity inventory, and surface form are otherwise identical.

\subsection{Metric and Intervention Definitions}\label{app:metrics}

This appendix collects the explicit formulas, cohort usage, and intervention protocols used in the analyses of \S\ref{sec:rq1}--\S\ref{sec:rq3}.

\paragraph{Joint-label coding.}
For each original/perturbed pair $i$, we map the extracted answer on both inputs into $\{A,B,O\}$, where $A$ is the original oracle answer $A_i$, $B$ is the perturbed oracle answer $B_i$, and $O$ is any other extracted answer. The resulting two-letter code $\ell_t^M(i)$ includes \textbf{AB}, the counterfactually adaptive cell; \textbf{AA}, original-answer retention; \textbf{AO}, originally correct but predicts neither $A_i$ nor $B_i$ on the perturbed input; \textbf{BA}, the reversed pattern; and \textbf{BB}, predicting the perturbed oracle on both inputs. Cells with $O$ in either coordinate are retained in the full $3 \times 3$ table and grouped as needed in figures.

\paragraph{Contrast Consistency, Counterfactual Responsiveness, and complementary rates.}
For each model checkpoint $M_t$ and oracle pair $i$, let $A_i$ be the original oracle answer and $B_i$ be the perturbed oracle answer. We report original-question accuracy
\[
\mathrm{acc}_t^M = \Pr[\hat{y}_{M_t}(x_i) = A_i],
\]
Contrast Consistency in the sense of~\citet{gardner-etal-2020-evaluating}'s contrast sets,
\[
\mathrm{CC}(M_t) = \Pr\!\left[\hat{y}_{M_t}(x_i) = A_i \,\land\, \hat{y}_{M_t}(x'_i) = B_i\right],
\]
and counterfactual responsiveness (CFR), the conditional form of $\mathrm{CC}$ in the lineage of \citet{kaushik2020_counterfactual,lanham2023_measuring_faithfulness},
\[
\begin{aligned}
\mathrm{CFR}(M_t)
&= \Pr\!\left[\hat{y}_{M_t}(x'_i) = B_i \,\middle|\, \hat{y}_{M_t}(x_i) = A_i\right] \\
&= \mathrm{CC}(M_t) / \mathrm{acc}_t^M.
\end{aligned}
\]
Equivalently, $\mathrm{CC}$ is the AB rate on the full $405$-pair denominator and $\mathrm{CFR}$ is the AB rate among the originally-correct pairs (AB$+$AA$+$AO); the complementary AA (retention) and AO fractions on that same denominator account for the rest. We report $\mathrm{CC}$ and $\mathrm{CFR}$ jointly because $\mathrm{CC}$ mixes capability with responsiveness, whereas $\mathrm{CFR}$ conditions counterfactual responsiveness on originally correct predictions. High $\mathrm{CFR}$ records oracle following but not how the answer was produced; it may also reflect a shortcut or computation that bypasses the latent reasoning states.


\paragraph{Transition cohorts on ProsQA $\times$ COCONUT.}
\begin{table}[ht]
\centering
\small
\begin{tabular}{lcccc}
\toprule
Cohort & $n$ & ep~$15$ & ep~$16$ & ep~$48$ \\
\midrule
$G_{\text{stable}}$ & 41 & $\mathrm{AB}$ & $-$ & $\mathrm{AB}$ \\
$G_{\text{drift}}$  & 68 & $\mathrm{AB}$ & $\mathrm{AA}$ & $\mathrm{AA}$ \\
$G_{\text{retain}}$ & 37 & $\mathrm{AA}$ & $-$ & $\mathrm{AA}$ \\
\bottomrule
\end{tabular}
\caption{Cohort criteria for the COCONUT ProsQA analysis. Epochs~15 and 16 capture the rapid middle-training change, and epoch~48 is validation-selected. A $-$ entry means that epoch~16 is not part of that cohort definition.}
\label{tab:rq2-cohort-def}
\end{table}

We use the three COCONUT ProsQA cohorts defined in Table~\ref{tab:rq2-cohort-def}. Epoch~$16$ is reported for every cohort, but for $G_{\text{stable}}$ and $G_{\text{retain}}$ it is not used to select examples; it is only an evaluation checkpoint for the intervention. Only $G_{\text{drift}}$ is conditioned on epoch~$16$, because it isolates examples that move from $\mathrm{AB}$ at epoch~$15$ to $\mathrm{AA}$ at the curriculum boundary and are also $\mathrm{AA}$ at epoch~$48$. The resulting cohorts contain $41$ stable, $68$ drift and $37$ retention pairs.
On each cohort we apply the all-step \texttt{norm-noise} intervention at epochs~$15$, $16$, and $48$, using five noise trials per pair. For a pair whose unpatched prediction is one of the two candidates, let $b_i$ denote that prediction and $o_i$ the other candidate. We teacher-force each complete answer candidate after the baseline-generated answer prefix, held fixed across interventions within that checkpoint. Candidate log probabilities are summed over their tokens. Here, the superscript $\mathrm{clean}$ denotes the unpatched computation on the evaluated input, which may be original or edited. We measure the decrease in the unpatched prediction's margin:
\begin{align*}
d_i &= \left[\log p^{\text{clean}}(b_i)-\log p^{\text{clean}}(o_i)\right] \\
&\quad - \frac{1}{5}\sum_{r=1}^{5}\left[\log p^{\text{patch},r}(b_i)-\log p^{\text{patch},r}(o_i)\right].
\end{align*}
Positive $d_i$ means that patching weakens the unpatched binary prediction; it does not necessarily shift probability toward the perturbed answer $B_i$. Table~\ref{tab:rq2-cohort} reports the median $d_i$ across pairs with a finite margin. In $G_{\text{drift}}$, the same example is solved correctly under perturbation at epoch~$15$ but not at epochs~$16$ and $48$, so comparisons across these checkpoints do not change item composition.

\paragraph{Activation patching protocol.}
For the activation trajectories, each checkpoint and intervention uses four $256$-question subsamples indexed by $\{0,1,2,3\}$. ProsQA subsamples are drawn from the original $500$ test questions; the paired analyses use the $405$ accepted counterfactual pairs. We compare a clean forward pass with a patched forward pass on each subset. The all-step intervention replaces every latent position in the current curriculum stage. We use \texttt{zero} replacement, corresponding-position \texttt{mean} replacement computed from the same $256$-question subset, and \texttt{norm-noise} replacement with isotropic Gaussian noise rescaled to the original latent norm. Norm-noise is repeated for three trials per question. For COCONUT, replacement is applied to the continuous-thought embedding before it is fed back as the next input embedding. For CODI, replacement is applied to the post-projection hidden state used as the next latent-loop input. All three task formulations use greedy generation with a $64$-token maximum. Answers are extracted with the task-specific parsers used by the trajectory evaluation.

\paragraph{Output-level metrics for activation patching.}
The output change rate is
\[
\mathrm{OCR} = \Pr[\hat{y}^{\text{patch}} \neq \hat{y}^{\text{clean}}],
\]
the preserved-when-correct rate is
\[
\mathrm{PWC} = \Pr[\hat{y}^{\text{patch}} = y \mid \hat{y}^{\text{clean}} = y],
\]
where $y$ is the task gold answer after the task-specific answer normalisation. The flip decomposition refines output changes by correctness direction:
\begin{align*}
\Pr[\mathrm{c}\to\mathrm{w}]
&= \Pr[\hat{y}^{\text{patch}} \neq y \mid \hat{y}^{\text{clean}} = y], \\
\Pr[\mathrm{w}\to\mathrm{c}]
&= \Pr[\hat{y}^{\text{patch}} = y \mid \hat{y}^{\text{clean}} \neq y],
\end{align*}
These rates distinguish harmful changes from corrections. Their denominators are the initially correct and initially wrong predictions, respectively. For GSM-open, $\mathrm{OCR}$ is computed after the same numeric-answer normalisation used for accuracy.

\paragraph{Sampling and uncertainty.}
The activation trajectory estimates use $1{,}024$ baseline question evaluations per checkpoint and $3{,}072$ patched evaluations for \texttt{norm-noise}. Questions may overlap across the four subsamples. Curves show the mean and one standard deviation across subsamples. Cohort analyses use five noise trials per input, while the additional controls use the sample sizes and repetitions specified in Appendix~\ref{app:controls}.

\subsection{Additional Results}\label{app:additional-results}

\begin{figure}[ht]
\centering
\includegraphics[width=\linewidth]{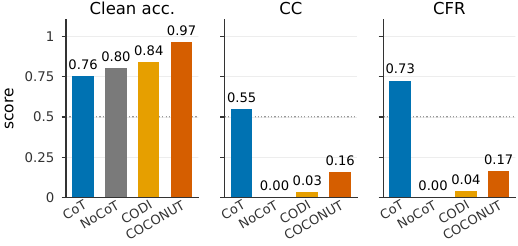}
\caption{Clean accuracy, Contrast Consistency ($\mathrm{CC}$), and counterfactual responsiveness ($\mathrm{CFR}$) on the 405 ProsQA pairs at each model's validation-selected checkpoint: CoT epoch~30, NoCoT epoch~14, CODI epoch~37, and COCONUT epoch~48.}
\label{fig:rq1-clean-vs-cfr}
\end{figure}

\paragraph{COCONUT curriculum analysis.}
We resample the same $405$ question IDs $10{,}000$ times and recompute CFR with its conditional denominator. Table~\ref{tab:coconut-curriculum-audit} reports all six transitions. CFR decreases at several points, with the largest decrease occurring within stages~2--3 between epochs~15 and 16. This analysis quantifies the timing and size of the changes along the training trajectory.

\begin{table}[h]
\centering
\footnotesize
\setlength{\tabcolsep}{0pt}
\begin{tabular}{@{}l@{\hspace{1.5pt}}c@{\hspace{1.5pt}}c@{\hspace{1.5pt}}l@{\hspace{4.5pt}}r@{}}
\toprule
\multicolumn{5}{@{}l}{\textit{All curriculum transitions}} \\
\midrule
Trans. & Latent & $\Delta$Acc. & $\Delta$CFR [95\% CI] & $p_{\mathrm{H}}$ \\
\midrule
$5\to6$ & $0\to1$ & $+.0790$ & $-.0770$ [$-.143,-.011$] & $.0711$ \\
$10\to11$ & $1\to2$ & $+.0494$ & $-.0251$ [$-.088,+.039$] & $.4548$ \\
\textbf{$15\to16$} & \textbf{$2\to3$} & \textbf{$+.0420$} & \textbf{$-.3056$ [$-.371,-.238$]} & \textbf{$.0006$} \\
$20\to21$ & $3\to4$ & $+.0123$ & $-.1651$ [$-.215,-.118$] & $.0006$ \\
$25\to26$ & $4\to5$ & $+.0123$ & $-.0770$ [$-.115,-.040$] & $.0006$ \\
$30\to31$ & $5\to6$ & $-.0148$ & $-.0306$ [$-.071,+.009$] & $.2684$ \\
\midrule
\multicolumn{5}{@{}l}{\textit{Epoch $15\to16$ compared with the other transitions}} \\
\midrule
Comparator $b$ & \multicolumn{4}{c}{$\Delta\mathrm{CFR}_{15\to16}-\Delta\mathrm{CFR}_{b}$ [95\% CI]} \\
\midrule
$5\to6$ & \multicolumn{4}{c}{$-.2287$ [$-.326,-.133$]} \\
$10\to11$ & \multicolumn{4}{c}{$-.2806$ [$-.373,-.190$]} \\
$20\to21$ & \multicolumn{4}{c}{$-.1405$ [$-.225,-.054$]} \\
$25\to26$ & \multicolumn{4}{c}{$-.2287$ [$-.304,-.154$]} \\
$30\to31$ & \multicolumn{4}{c}{$-.2751$ [$-.352,-.196$]} \\
\bottomrule
\end{tabular}
\caption{COCONUT curriculum analysis. Changes are computed as the value after a transition minus the value before it. The first panel applies Holm correction across the six CFR tests. In the second panel, negative values indicate a larger CFR decrease at epoch~$15\to16$; all five comparisons have Holm-adjusted $p\leq 0.0009$. Intervals reflect paired-example resampling within this training run. Accuracy intervals are exploratory and omitted for space.}
\label{tab:coconut-curriculum-audit}
\end{table}

For the Llama-1B stages 2--3 interval, we apply the same paired bootstrap to epochs~6 and 9. CFR decreases by $42.3$ percentage points. The $95\%$ confidence interval for the final-minus-initial change is $[-47.9,-36.7]$ percentage points.

\begin{figure}[ht]
\centering
\includegraphics[width=\linewidth]{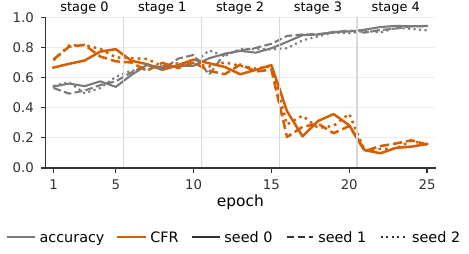}
\caption{Accuracy and CFR of GPT-2 COCONUT on ProsQA for three training seeds through epoch~25. Seed~0 is the run analyzed in the main text; seeds~1 and~2 repeat its settings and schedule.}
\label{fig:seeds}
\end{figure}

\paragraph{Additional training seeds.}
Two further GPT-2 COCONUT runs repeat the original settings and schedule with seeds~1 and~2 through epoch~25, the end of stage~4. Figure~\ref{fig:seeds} shows accuracy and CFR for the three seeds. Across the seeds, mean CFR lies between $0.65$ and $0.68$ in stage~2, between $0.26$ and $0.31$ in stage~3, and between $0.13$ and $0.15$ in stage~4, while mean accuracy rises from about $0.77$ to about $0.88$ and then $0.93$. From epoch~15 to epoch~16, CFR falls by $0.31$, $0.45$ and $0.39$. The three seeds also repeat the replacement analysis of \S\ref{sec:rq2}, and Table~\ref{tab:seed-patching} reports \texttt{norm-noise} $\mathrm{OCR}$ and $\mathrm{PWC}$ at six checkpoints. In every seed, OCR declines across the curriculum, from between $11\%$ and $14\%$ at epoch~6 to between $2\%$ and $4\%$ at epoch~20, and PWC rises above $98\%$. At these checkpoints, OCR stays below $10\%$ under \texttt{zero} replacement and at most $1\%$ under \texttt{mean} replacement.

\begin{table}[ht]
\centering
\small
\setlength{\tabcolsep}{4pt}
\begin{tabular*}{\linewidth}{@{\extracolsep{\fill}}rcccccc@{}}
\toprule
 & \multicolumn{3}{c}{OCR (\%)} & \multicolumn{3}{c}{PWC (\%)} \\
\cmidrule(lr){2-4}\cmidrule(lr){5-7}
Epoch & seed 0 & seed 1 & seed 2 & seed 0 & seed 1 & seed 2 \\
\midrule
6 & 13.9 & 11.3 & 12.9 & 94.7 & 96.2 & 95.9 \\
10 & 10.1 & 7.5 & 9.0 & 96.2 & 97.2 & 97.6 \\
15 & 6.5 & 6.4 & 5.9 & 96.8 & 97.6 & 97.6 \\
16 & 6.2 & 7.0 & 4.9 & 97.6 & 95.8 & 98.1 \\
20 & 3.8 & 3.4 & 2.2 & 98.4 & 98.2 & 99.6 \\
25 & -- & 2.0 & 1.1 & -- & 99.1 & 99.8 \\
\bottomrule
\end{tabular*}
\caption{\texttt{Norm-noise} replacement of all latent states for three training seeds of GPT-2 COCONUT on ProsQA. Values are means over four 256-question evaluation subsamples. Seed~0 has no evaluation at epoch~25.}
\label{tab:seed-patching}
\end{table}

\paragraph{Stage transitions versus optimizer resets.}
COCONUT recreates AdamW at every one of the $49$ epoch boundaries, whereas the latent stage advances at six of them. Across the $43$ within-stage boundaries, the change in CFR has mean $+0.004$ and standard deviation $0.044$; the largest within-stage decrease is $-0.167$ at epoch~$16\to17$, immediately after a transition. At the six stage transitions the change is $-0.077$, $-0.025$, $-0.306$, $-0.165$, $-0.077$ and $-0.031$. Optimizer resets alone therefore do not produce the decreases, although this comparison cannot exclude an interaction between resets and the curriculum.

\paragraph{Consistency across evaluation subsamples.}
Table~\ref{tab:subsample-endpoints} summarizes the first and last evaluated checkpoints for GPT-2 COCONUT under \texttt{norm-noise}. All four subsamples preserve the direction of each trajectory.

\begin{table}[ht]
\centering
\small
\begin{tabular}{lcc}
\toprule
Setting (epochs) & First OCR & Last OCR \\
\midrule
ProsQA (6--48) & $0.139\pm0.009$ & $0.000\pm0.000$ \\
GSM-Choice (4--25) & $0.316\pm0.013$ & $0.101\pm0.007$ \\
GSM-open (4--25) & $0.641\pm0.011$ & $0.874\pm0.017$ \\
\bottomrule
\end{tabular}
\caption{OCR mean and standard deviation across four evaluation subsamples at the trajectory endpoints.}
\label{tab:subsample-endpoints}
\end{table}

Table~\ref{tab:rq2-coconut-patterns} lists the core joint label counts for COCONUT on ProsQA across selected checkpoints. The counts show the movement from counterfactual following to original-answer retention during the middle curriculum stages.

\begin{table}[h]
\centering
\small
\begin{tabular}{rccccc}
\toprule
Epoch & $\mathrm{AB}$ & $\mathrm{AA}$ & $\mathrm{BA}$ & $\mathrm{BB}$ & Clean acc \\
\midrule
5 & 172 & 16 & 0 & 6 & 0.538 \\
10 & 198 & 42 & 3 & 14 & 0.679 \\
15 & \textbf{220} & 41 & 0 & 8 & 0.795 \\
16 & 128 & \textbf{157} & 16 & 22 & 0.837 \\
17 & 75 & 259 & 18 & 14 & 0.881 \\
20 & 104 & 240 & 13 & 9 & 0.906 \\
30 & 78 & 312 & 5 & 8 & 0.968 \\
48 (selected) & 65 & 326 & 5 & 9 & 0.965 \\
\bottomrule
\end{tabular}
\caption{COCONUT on ProsQA: joint label counts on the $405$ oracle pairs across selected checkpoints. Counterfactual following moves toward original-answer retention during the middle curriculum stages, while clean accuracy rises.}
\label{tab:rq2-coconut-patterns}
\end{table}

\paragraph{Flip decomposition and accuracy on GSM.}
On GSM-Choice, \texttt{zero} replacement preserves $91.9\%$ of initially correct predictions and corrects $18.0\%$ of initially wrong predictions at the validation-selected checkpoint (Table~\ref{tab:rq4-gsm-choice-flips}). For the corrected examples, the model answers incorrectly with its original latent states but correctly after those states are replaced by zeros, with the question and decoding procedure unchanged. The original latent states can therefore sometimes interfere with correct prediction. Table~\ref{tab:rq4-gsm-format-ranking} lists the validation accuracy of COCONUT and CoT on both formulations.

\begin{table}[!ht]
\centering
\small
\begin{tabular}{rlccc}
\toprule
Epoch & Patch & PWC & c$\to$w & w$\to$c \\
\midrule
4  & zero       & 0.801 & 0.199 & 0.510 \\
13 & zero       & 0.850 & 0.150 & 0.280 \\
25 & zero       & 0.919 & 0.081 & 0.180 \\
25 & mean       & 0.933 & 0.067 & 0.155 \\
25 & norm-noise & 0.922 & 0.078 & 0.162 \\
\bottomrule
\end{tabular}
\caption{Flip decomposition on GSM-Choice $\times$ COCONUT, averaged across four evaluation subsamples. PWC and c$\to$w condition on initially correct predictions; w$\to$c conditions on initially wrong predictions.}
\label{tab:rq4-gsm-choice-flips}
\end{table}

\begin{table}[!ht]
\centering
\small
\begin{tabular}{lcc}
\toprule
 & GSM-Choice & GSM-open \\
\midrule
COCONUT (selected ep.) & \textbf{0.868} & 0.356 \\
CoT     (selected ep.) & 0.828 & \textbf{0.446} \\
\midrule
$\Delta$ (COCONUT $-$ CoT) & $+0.040$ & $-0.090$ \\
\bottomrule
\end{tabular}
\caption{Validation accuracy at each model's validation-selected epoch on the same GSM8K problems in two answer formulations.}
\label{tab:rq4-gsm-format-ranking}
\end{table}





\subsection{Additional Edit and Intervention Controls}\label{app:controls}

\paragraph{Edits that preserve or change the answer.}
This control tests whether models fail broadly under premise edits. We construct a single-premise edit that preserves the BFS answer for $345$ of the $405$ original pairs and compare both edit types on these matched examples. Among the preserving edits, $286$ also preserve the gold shortest-path distance. Table~\ref{tab:edit-controls} reports accuracy across checkpoints. Accuracy rises for preserving edits and falls for changing edits. Correct predictions under preserving edits can result from ignoring the edit, so the contrast remains compatible with difficulty generalizing to answer-changing edits.

\begin{table}[ht]
\centering
\small
\begin{tabular}{lrrr}
\toprule
Edit & ep 15 & ep 16 & ep 48 \\
\midrule
Answer preserved & 79.42 & 86.67 & 96.81 \\
Answer changed & 62.61 & 43.19 & 20.87 \\
\bottomrule
\end{tabular}
\caption{Accuracy (\%) on the same $345$ examples. These denominators differ from the full $405$-pair trajectory.}
\label{tab:edit-controls}
\end{table}

\paragraph{Graph validity.}
The edits test the same ProsQA reachability task. BFS verifies for every pair that the original answer becomes unreachable and the alternative reachable. Each edited graph can also be generated by Algorithm~1 of \citet{hao2025_COCONUT}, allowing node reordering, which supports task validity. A change to the graph may still affect generalization. Unconditional edited accuracy falls together with CFR in \S\ref{sec:rq1}, so the decline in CFR is not an effect of its denominator.

\paragraph{Fixed intervention scope on ProsQA.}
To keep the perturbation scope constant as latent stages grow, we replace position~0 or the first two positions with norm-matched noise. We evaluate all $405$ original/edited pairs at the checkpoints in Table~\ref{tab:fixed-scope-controls}, using three trials per input. OCR declines by late training under both scopes.

\begin{table}[ht]
\centering
\small
\begin{tabular}{llrrr}
\toprule
Input & Positions & ep 15 & ep 16 & ep 48 \\
\midrule
Original & $\{0\}$ & 3.95 & 2.96 & 0.00 \\
Edited & $\{0\}$ & 7.90 & 6.75 & 0.49 \\
Original & $\{0,1\}$ & 5.35 & 4.77 & 0.00 \\
Edited & $\{0,1\}$ & 10.62 & 9.79 & 0.66 \\
\bottomrule
\end{tabular}
\caption{OCR (\%) under fixed-position norm-noise on ProsQA. Trials are averaged within questions.}
\label{tab:fixed-scope-controls}
\end{table}

\paragraph{Transfer between original and edited inputs.}
We test whether latent states from the paired input change the recipient's answer. States are transferred in both directions at epochs~15, 16 and~48, replacing either the first two or all current latent positions. Each scope changes only two of $2{,}430$ greedy answers. The transfer therefore has little behavioral effect under this setup. The recipient retains its own question and can recover information through other computation, leaving the states' reasoning content unresolved. Self-patching reproduces baseline tokens, and vector checks verify the replacements.

\paragraph{Margin readout and cohort uncertainty.}
The primary analysis holds the baseline answer prefix fixed within each checkpoint. We include examples whose baseline answers identify a candidate and yield a finite oriented margin. Table~\ref{tab:margin-readout-details} lists the eligible counts. Checkpoint differences use the same eligible example IDs and $4{,}000$ paired-question bootstrap draws. For the drift cohort's edited inputs, the fixed-prefix margin decrease falls by $1.12$ nats from epoch~15 to~16. The $95\%$ interval for this change is $[-1.57,-0.14]$ nats. These estimates describe the selected cohorts within this training run; the stable cohort also shows a decline.

\begin{table}[ht]
\centering
\small
\begin{tabular}{lrrr}
\toprule
 & ep 15 & ep 16 & ep 48 \\
\midrule
Stable & 41 & 35 & 41 \\
Drift, each input & 68 & 68 & 68 \\
Retention & 37 & 29 & 37 \\
\bottomrule
\end{tabular}
\caption{Examples in each cohort that are eligible for the fixed-prefix margins of Table~\ref{tab:rq2-cohort}.}
\label{tab:margin-readout-details}
\end{table}

\paragraph{GSM replacement across questions.}
We replace all latent states with states generated for another question in the same setting and checkpoint. Each of the eight checkpoints uses the same $256$ questions and three random donor pairings, excluding self-donors. Greedy decoding and the $64$-token limit remain unchanged. All baseline outputs match the archived evaluation, and self-patching reproduces baseline tokens.

\begin{table}[ht]
\centering
\small
\begin{tabular}{lrr}
\toprule
Setting & ep 4 & ep 25 \\
\midrule
Open & 88.67 & 93.23 \\
Choice & 33.33 & 11.33 \\
\bottomrule
\end{tabular}
\caption{OCR (\%) when all latent states are replaced with those of another question.}
\label{tab:gsm-donor-controls}
\end{table}

Table~\ref{tab:gsm-donor-controls} shows opposite endpoint directions: OCR rises on Open and falls on Choice. The control tests sensitivity to states from other questions; interpreting these changes still requires accounting for output spaces and initializations. Pairings and bootstrap intervals quantify evaluation uncertainty within each training run.

\subsection{Diagnosis-Guided Training Runs}\label{app:rq4}

\paragraph{Runs and schedules.}
Unless noted, runs in \S\ref{sec:rq4} load checkpoint~15 of the GPT-2 COCONUT run and keep its batch size, learning rate, FP32 precision, seed and per-epoch optimizer recreation. The delayed-transition run trains checkpoints~16--21 at stage~2 and checkpoints~22--25 at stage~3. The compressed-curriculum runs train checkpoints~16--17 at stage~3, 18--19 at stage~4, 20--21 at stage~5 and 22--23 at stage~6. Each run takes between five and eight GPU-hours on one RTX PRO 6000.

\paragraph{Latent supervision.}
Consider an example with $n$ reasoning steps, where step~$j$ has the form ``Every $u_j$ is a $v_j$.''. At stage~$k$, COCONUT replaces the first $m=\min(k,n)$ steps with latent positions and trains on the set $T$ of remaining text tokens. Let $h_j$ be the hidden state that is fed back as the input of latent position~$j$, and let $c_j$ be the first token of $v_j$. Latent supervision adds one target per replaced step:
\begin{align*}
\mathcal{L} = -\frac{1}{|T|+m}\Bigl[&\sum_{t\in T}\log p_\theta(x_t\mid x_{<t})\\
&+ \sum_{j=1}^{m}\log p_\theta(c_j\mid h_j)\Bigr],
\end{align*}
where $p_\theta(\cdot\mid h_j)$ applies the language-model head to $h_j$. Without the second sum and with denominator $|T|$, this is the standard COCONUT loss. The change is implemented by placing the label $c_j$ at latent position~$j$; the model code and inference are unchanged. Latent positions beyond the number of steps of an example carry no label, and supervised positions have the same weight as text tokens. In a sample of $3{,}000$ training questions, $17\%$ of step targets share their first token with another concept in the same question.

\paragraph{Stage replay.}
At stage~$k$, the stage $k'$ used to train an example with $n$ steps is drawn independently in every epoch:
\begin{equation*}
k' =
\begin{cases}
k & \text{with probability } 1-\rho,\\
u & \text{with probability } \rho,
\end{cases}
\end{equation*}
where $u$ is uniform on $\{0,\ldots,\min(k,n)-1\}$ and $\rho=0.3$. In expectation, $30\%$ of the examples in an epoch are therefore replayed, and a replayed example keeps at least one explicit step because $k'<n$. In the runs that branch from epoch~5, $u$ instead takes the value $s$ with probability proportional to $\delta^{\min(k,n)-1-s}$, with $\delta=0.5$, so that the most recent earlier stage is the most likely one.

\paragraph{Results.}
In Figure~\ref{fig:rq4-steps} of the main text, chains of a given length lose responsiveness at the stage where they become fully latent or one stage earlier, whereas CoT stays near $0.7$ for every length throughout training. In the delayed-transition run, CFR stays between $0.64$ and $0.69$ across the six added stage-2 checkpoints and falls from $0.65$ to $0.44$ at the first stage-3 checkpoint. The $95\%$ interval for this change is $-0.29$ to $-0.15$, and three-step examples fall from $0.63$ to $0.25$.

\begin{figure}[ht]
\centering
\includegraphics[width=0.8\linewidth]{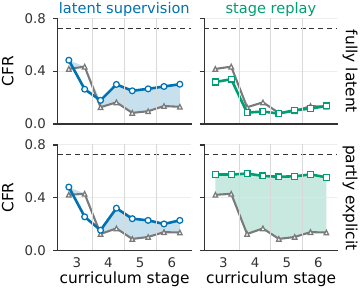}
\caption{CFR at every checkpoint of the runs that branch from epoch~15, drawn as in the grid of Figure~\ref{fig:rq4-intervention}. Grey is the standard run of this branch, the shaded area is the gain, and the dashed line is CoT.}
\label{fig:rq4-branch15}
\end{figure}

Table~\ref{tab:rq4-summary} reports accuracy, CFR and CC at the final curriculum stage of every run. CC, which does not condition on the original prediction, leads to the same conclusions as CFR. Intervals are $95\%$ percentile intervals from $4{,}000$ bootstrap draws that resample the $405$ pairs, with the same draw applied to both runs. The interval excludes zero for every gain except stage replay under fully latent inference. Figure~\ref{fig:rq4-branch15} shows the runs that branch from epoch~15 at every checkpoint, and Figure~\ref{fig:rq4-intervention} those that branch from epoch~5.

\paragraph{Runs from epoch 5.}
Four runs load checkpoint~5, the last checkpoint trained with all steps explicit, and train stages~1--6 for two checkpoints each: a standard run, latent supervision, stage replay, and a combined run that applies latent supervision and the same replay. The standard and latent-supervision runs are close at the first stage-3 checkpoint, and the supervised run has higher CC and accuracy at every checkpoint from stage~4 onward.

\begin{table*}[ht]
\centering
\small
\setlength{\tabcolsep}{5pt}
\begin{tabular}{lcccc@{\hskip 16pt}cccc}
\toprule
 & \multicolumn{4}{c}{Fully latent inference} & \multicolumn{4}{c}{Partly explicit inference} \\
\cmidrule(r{16pt}){2-5}\cmidrule(lr){6-9}
Training & Acc & CFR & Gain in CFR [95\% CI] & CC & Acc & CFR & Gain in CFR [95\% CI] & CC \\
\midrule
\multicolumn{9}{l}{\emph{Branch at epoch 15}} \\
Standard & .941 & .133 &  & .125 & .941 & .136 &  & .128 \\
Latent supervision & .967 & \textbf{.292} & $+$.160 [.120, .200] & .283 & .867 & .214 & $+$.077 [.036, .118] & .185 \\
Stage replay & .911 & .126 & $-$.007 [$-$.035, .020] & .115 & .812 & \textbf{.565} & $+$.429 [.381, .475] & .459 \\
\addlinespace[3pt]
\multicolumn{9}{l}{\emph{Branch at epoch 5}} \\
Standard & .896 & .101 &  & .090 & .895 & .101 &  & .090 \\
Latent supervision & .956 & \textbf{.296} & $+$.195 [.154, .237] & .283 & .846 & .274 & $+$.174 [.124, .224] & .232 \\
Stage replay & .917 & .093 & $-$.008 [$-$.037, .021] & .085 & .836 & .532 & $+$.431 [.384, .478] & .444 \\
Replay + supervision & .958 & .282 & $+$.182 [.139, .223] & .270 & .880 & \textbf{.555} & $+$.455 [.404, .504] & .489 \\
\midrule
CoT (reference) & .756 & .725 & & .548 & -- & -- & & -- \\
NoCoT (reference) & .802 & .003 & & .002 & -- & -- & & -- \\
\bottomrule
\end{tabular}
\caption{Accuracy, CFR and CC at the final curriculum stage, pooled over its two checkpoints. The gain is the difference in CFR from the standard run of the same branch under the same inference, with a paired-bootstrap interval. Fully latent inference uses all six latent positions; partly explicit inference uses two and generates later steps as text. CoT and NoCoT use none and are shown at their validation-selected checkpoints.}
\label{tab:rq4-summary}
\end{table*}

\subsection{The Use of Large Language Models}\label{app:ai-assistants}

We used LLMs for limited writing support, including refining grammar and improving language clarity and fluency. The scientific content, experimental design, analysis, claims, and citations were produced and verified by the authors. The authors reviewed and edited all LLM-generated content and take full responsibility for the final submission.

\subsection{Notation and Terminology}\label{app:notation}

Table~\ref{tab:notation} collects the symbols and terms used across the paper.

\clearpage
\begin{table*}[!t]
\centering
\small
\begin{tabular}{p{0.20\textwidth}p{0.72\textwidth}}
\toprule
Symbol / Term & Meaning \\
\midrule
\multicolumn{2}{l}{\emph{Training paradigms}} \\
CoT & Chain-of-thought baseline: generates explicit step-by-step text tokens before the answer. \\
NoCoT & Answer-only baseline: trained to emit the answer directly with no intermediate text. \\
COCONUT & Latent-reasoning method that replaces explicit reasoning tokens with continuous thoughts via a curriculum~\citep{hao2025_COCONUT}. \\
CODI & Latent-reasoning method that compresses CoT into latent thoughts via self-distillation~\citep{shen2025_codi}. \\
\midrule
\multicolumn{2}{l}{\emph{Datasets}} \\
ProsQA & Reachability QA over small directed graphs; each question has a binary candidate answer. \\
GSM-open & Open-ended numerical answer format of GSM8K~\citep{cobbe2021_gsm8k}. \\
GSM-Choice & Two-option multiple-choice reformulation of GSM8K aligned with the binary interface of ProsQA. \\
\midrule
\multicolumn{2}{l}{\emph{Pairs, inputs, and oracle answers}} \\
$x_i$, $x'_i$ & Original and perturbed input for pair $i$. \\
$A_i$, $B_i$ & Original oracle answer and perturbed oracle answer. \\
$t^-$ & Negative target node used to construct the perturbation. \\
$\hat{y}_{M_t}(\cdot)$ & Predicted answer of checkpoint $M_t$ on a given input. \\
\midrule
\multicolumn{2}{l}{\emph{Joint labels (per pair)}} \\
$\ell_t^M(i) \in \{A,B,O\}^2$ & Joint label encoding (original-answer, perturbed-answer) for checkpoint $M_t$ on pair $i$, where $O$ is any answer outside $\{A_i, B_i\}$. \\
$\mathrm{AB}$ & Original $\to A$, perturbed $\to B$; counterfactually adaptive. \\
$\mathrm{AA}$ & Original $\to A$, perturbed $\to A$; original-answer retention. \\
$\mathrm{AO}$ & Original $\to A$, perturbed $\to O$; correct on the original, off-target on the perturbed input. \\
\midrule
\multicolumn{2}{l}{\emph{Behavioural metrics}} \\
$\mathrm{acc}_t^M$ & Original-question accuracy at checkpoint $M_t$. \\
$\mathrm{CC}(M_t)$ & Contrast Consistency, $\Pr[\hat{y}(x_i)=A_i \land \hat{y}(x'_i)=B_i]$~\citep{gardner-etal-2020-evaluating}. \\
$\mathrm{CFR}(M_t)$ & Counterfactual responsiveness, $\Pr[\hat{y}(x'_i)=B_i \mid \hat{y}(x_i)=A_i]$. \\
\midrule
\multicolumn{2}{l}{\emph{Activation-level metrics}} \\
$\mathrm{OCR}$ & Output change rate, $\Pr[\hat{y}^{\text{patch}} \neq \hat{y}^{\text{clean}}]$. \\
$\mathrm{PWC}$ & Preserved-when-correct rate, $\Pr[\hat{y}^{\text{patch}}=y \mid \hat{y}^{\text{clean}}=y]$. \\
$\mathrm{acc}_c$, $\mathrm{acc}_p$ & Clean and patched accuracy on the same evaluation set. \\
$\Pr[\mathrm{c}\to\mathrm{w}]$ & Conditional rate of becoming wrong after patching among initially correct predictions. \\
$\Pr[\mathrm{w}\to\mathrm{c}]$ & Conditional rate of becoming correct after patching among initially wrong predictions. \\
$d_i$ & Decrease in the unpatched binary prediction's log-probability margin after patching. Positive values mean that prediction is weakened. \\
\midrule
\multicolumn{2}{l}{\emph{Interventions}} \\
\texttt{zero} & Replace the latent activation at the target position with the zero vector. \\
\texttt{mean} & Replace with the corresponding-position mean activation from the same evaluation subsample. \\
\texttt{norm-noise} & Replace with isotropic Gaussian noise rescaled to the original latent norm. \\
\midrule
\multicolumn{2}{l}{\emph{Cohorts (ProsQA $\times$ COCONUT)}} \\
$G_{\text{stable}}$ & Pairs with $\ell_{15}=\ell_{48}=\mathrm{AB}$ ($n=41$); counterfactually adaptive at epochs~$15$ and $48$. \\
$G_{\text{drift}}$ & Pairs with $\ell_{15}=\mathrm{AB}$, $\ell_{16}=\mathrm{AA}$, $\ell_{48}=\mathrm{AA}$ ($n=68$); adaptive at epoch~$15$ and retains the original answer at epochs~$16$ and $48$. \\
$G_{\text{retain}}$ & Pairs with $\ell_{15}=\ell_{48}=\mathrm{AA}$ ($n=37$); retains the original answer at epochs~$15$ and $48$. \\
\midrule
\multicolumn{2}{l}{\emph{Checkpoints and curriculum}} \\
$M_t$ & Model at training checkpoint $t$. \\
$\{M_t\}$ & Set of saved checkpoints evaluated for a training run. \\
$n_{\text{latent}}$ & Number of latent positions in the current curriculum stage. \\
$c_{\text{thought}}$ & COCONUT hyperparameter: continuous thoughts inserted per replaced reasoning step. \\
\bottomrule
\end{tabular}
\caption{Notation and terminology used throughout the paper.}
\label{tab:notation}
\end{table*}

\clearpage

\end{document}